\pdfoutput=1

\documentclass[11pt]{article}

\usepackage[final]{acl}

\usepackage{times}
\usepackage{latexsym}

\usepackage[T1]{fontenc}

\usepackage[utf8]{inputenc}

\usepackage{microtype}

\usepackage{inconsolata}

\usepackage{amssymb,amsmath}

\newenvironment{myquote}{\list{}{\leftmargin=0.1in\rightmargin=0.1in}\item[]}%
  {\endlist}

\usepackage{xcolor}

\usepackage{multirow}
\usepackage{tabularx}
\usepackage{booktabs}
\usepackage{xcolor,caption,subcaption,graphicx}

\title{Ontology Completion with Natural Language Inference and\\ Concept Embeddings: An Analysis}

\author{Na Li$^{3}$, Thomas Bailleux$^{1}$, Zied Bouraoui$^{1}$, Steven Schockaert$^{2}$\\
  $^{1}$ CRIL CNRS \& University of Artois, France \quad
  $^{2}$ CardiffNLP, Cardiff University, UK \\
  $^{3}$ University of Shanghai for Science and Technology, China\\
  \texttt{\{bailleux,bouraoui\}@cril.fr}   \quad \texttt{schockaerts1@cardiff.ac.uk} \\
   \texttt{li\_na@usst.edu.cn}\\}

\begin{document}
\maketitle
\begin{abstract}
We consider the problem of finding plausible knowledge that is missing from a given ontology, as a generalisation of the well-studied taxonomy expansion task. One line of work treats this task as a Natural Language Inference (NLI) problem, thus relying on the knowledge captured by language models to identify the missing knowledge. Another line of work uses concept embeddings to identify what different concepts have in common, taking inspiration from cognitive models for category based induction. These two approaches are intuitively complementary, but their effectiveness has not yet been compared. In this paper, we introduce a benchmark for evaluating ontology completion methods and thoroughly analyse the strengths and weaknesses of both approaches. We find that both approaches are indeed complementary, with hybrid strategies achieving the best overall results. We also find that the task is highly challenging for Large Language Models, even after fine-tuning.
\end{abstract}

\section{Introduction}
Ontologies, in the context of artificial intelligence, are essentially sets of rules which describe how the concepts from a given domain are related. They generalise taxonomies by expressing these relationships using logical connectives, which makes it possible to describe conceptual relationships in a more precise way. Throughout this paper, we will use the common description logic syntax for encoding ontology rules. For instance, an ontology might contain the following rules:
\begin{align}
\mathsf{Biologist} \sqcap (\exists\, \mathsf{livesIn}.\mathsf{UK}) &\sqsubseteq \mathsf{UKScientist}\label{eqExBritishScientist1}\\
\mathsf{Geologist} \sqcap (\exists\, \mathsf{livesIn}.\mathsf{UK}) &\sqsubseteq \mathsf{UKScientist}\label{eqExBritishScientist2}\\
\mathsf{Chemist} \sqcap (\exists\, \mathsf{livesIn}.\mathsf{UK}) &\sqsubseteq \mathsf{UKScientist}\label{eqExBritishScientist3}
\end{align}
In this syntax, rules are formulated as concept inclusions $X \sqsubseteq Y$, which encode that every instance of the concept $X$ is also an instance of the concept $Y$. The connective $\sqcap$ corresponds to intersection and $\exists r. C$ is the set of concepts that are related through the relation $r$ to some concept from $C$. For instance, \eqref{eqExBritishScientist1} expresses the knowledge that every scientist who is located somewhere in the UK is called a ``UK scientist''.\footnote{A detailed understanding of description logics, or their syntax, is not needed for this paper. However, the interested reader may refer to \citet{DBLP:series/ihis/BaaderHS09} for more details.}
We consider the problem of predicting missing rules in a given ontology. While this can be studied from multiple angles \cite{DBLP:journals/ki/Ozaki20}, we consider the setting where only the rules are given, meaning that we cannot learn rules from examples. We focus on methods that exploit the fact that the concepts involved are natural language terms, about which we have prior knowledge that can be exploited. Two main classes of methods have been studied for this setting.

First, we can simply treat the problem of predicting missing rules as a \emph{Natural Language Inference} (NLI) problem. Given a candidate rule such as \eqref{eqExBritishScientist1}, we then first need a verbalisation of the premise and hypothesis. For instance, the left-hand side of the concept inclusion might be translated to the premise ``a biologist who lives in the UK'' and the right-hand side might be translated to the hypothesis ``a UK Scientist''. While some care is needed about how the premise and hypothesis are formulated (e.g.\ how concept names referring the multi-word expressions are tokenised), this approach is intuitive and conceptually straightforward. \citet{DBLP:journals/www/ChenHGJDH23} implemented this strategy by fine-tuning a BERT \cite{devlin-etal-2019-bert} model on the rules from a given ontology. This strategy crucially relies on the knowledge which is captured by the pre-trained NLI model. However, this can be a problem for specialised ontologies, which may use concept names that do not refer to well-known natural language terms, or which use such terms with a meaning that is more specific than their standard meaning. 

The second approach relies on pre-trained \emph{concept embeddings}, e.g.\ obtained from standard word embeddings \cite{mikolov-etal-2013-linguistic,pennington-etal-2014-glove} or distilled from pre-trained language models \cite{DBLP:conf/ijcai/LiBCAGS21,liu-etal-2021-fast}. Essentially, pre-trained concept embeddings provide prior knowledge about the similarity structure of the concepts. To see why such knowledge can be useful for predicting missing rules, note that ontologies often contain large numbers of ``parallel rules'', i.e.\ concept inclusions which express essentially the same knowledge for a number of related concepts, as in the case of \eqref{eqExBritishScientist1}--\eqref{eqExBritishScientist3}. In the concept embedding space, concepts with a similar meaning will appear clustered together. This means in particular that the embedding of the concepts $\mathsf{Biologist}$, $\mathsf{Geologist}$ and $\mathsf{Chemist}$ will be similar. Based on this observation, for any concept $X$ whose embedding is also similar to that of $\mathsf{Biologist}$, $\mathsf{Geologist}$ and $\mathsf{Chemist}$, we can plausibly infer that there should be a counterpart to \eqref{eqExBritishScientist1}--\eqref{eqExBritishScientist3} for $X$. For instance, we would expect that the embedding of $\mathsf{Physicist}$ would also be similar and we can thus plausibly infer the following concept inclusion:
\begin{align}\label{eqRulePhysicistUK}
\mathsf{Physicist}\, {\sqcap}\, (\exists\, \mathsf{livesIn}.\mathsf{Britain}) \,{\sqsubseteq}\, \mathsf{UKScientist} 
\end{align}
Note that the justification for this inference comes purely from our prior knowledge about the relatedness of the concepts $\mathsf{Biologist}$, $\mathsf{Geologist}$, $\mathsf{Chemist}$ and $\mathsf{Physician}$. In particular, the concept $\mathsf{UKScientist}$ does not play any role in this process. This strategy can thus also be used if the concept in the head of the rule has a meaning which only makes sense within the context of the given ontology. However, its main drawback is that it can only be applied if suitable parallel rules are present. As such, we can expect this approach to be complementary to NLI based strategies. \citet{DBLP:conf/semweb/LiBS19} developed an embedding based method for ontology completion which uses a Graph Neural Network (GNN) to implement the aforementioned intuition.

While previous work has shown the feasibility of ontology completion, the strengths and weaknesses of existing methods are poorly understood. To address this, we make the following contributions:
\begin{itemize}
\item We introduce a benchmark for evaluating ontology completion methods. While previous work has focused on distinguishing held-out rules from randomly corrupted rules, our benchmark is specifically designed to include manually validated hard negatives. 
\item We thoroughly analyse the performance of NLI based approaches, comparing the BERT based models from \citet{DBLP:journals/www/ChenHGJDH23} with 
Large language models (LLMs). 
We similarly analyse the performance of concept embedding based methods, experimenting with different strategies for encoding the problem as a graph, different GNN architectures and different concept embeddings. 
\item We show that both types of methods are highly complementary. In particular, we find that simple hybrid strategies outperform any of the individual approaches.
\end{itemize}

\section{Related Work}
The problem of identifying plausible concept subsumptions is closely related to the problem of modelling hypernymy, which has been extensively studied, both as an intrinsic evaluation task for evaluating word representation models and within the context of applications such as taxonomy learning \cite{kozareva-hovy-2010-semi,bordea-etal-2016-semeval} and taxonomy enrichment \cite{jurgens-pilehvar-2016-semeval,takeoka-etal-2021-low}. While initial approaches were based on pattern matching in large text corpora \cite{hearst-1992-automatic}, later approaches have relied on word embeddings \cite{fu-etal-2014-learning,roller-etal-2014-inclusive} and subsequently on pre-trained language models \cite{chen-etal-2021-constructing,takeoka-etal-2021-low}. Essentially, the latter approaches fine-tune a BERT-based model \cite{devlin-etal-2019-bert} to classify a given word pair as either a positive or a negative instance of the hypernymy relation. Some approaches using GNNs have been proposed as well  \cite{DBLP:conf/www/ShenSX0W020,shang-etal-2020-taxonomy}. 

The problem of ontology completion has also been studied. \citet{DBLP:journals/corr/abs-2202-09791} fine-tune a BERT model to predict whether one concept subsumes another. In the basic version, this approach essentially follows the same strategy as BERT-based hypernym prediction approaches \cite{chen-etal-2021-constructing,takeoka-etal-2021-low}. However, they also analyse how some of the logical context of a concept can be included in the input to BERT. While this improves the results in some cases, these improvements are relatively small. This approach thus primarily relies on the ability of BERT to model hypernymy. As mentioned in the introduction, a radically different approach consists in using concept embeddings and relying on a GNN to model the ontology context \cite{DBLP:conf/semweb/LiBS19}. The idea of inferring plausible rules based on concept similarity has also been explored in a number of earlier works \cite{DBLP:conf/aaai/BouraouiS19,DBLP:journals/amai/dAmatoFFGL12,beltagy-etal-2016-representing}.
\citet{DBLP:journals/corr/abs-2302-06761} introduced ONTOLAMA, a benchmark for testing the ability of language models to recognise subsumption relations, including subsumption relations between complex concepts. This differs from the benchmark we introduce in this paper, as ONTOLAMA involves predicting the validity of a single rule, without any further ontology context. In contrast, we are specifically interested in methods that can take a set of existing rules into account.

In this paper, we focus on ontology completion methods that rely on the fact that NLP models (i.e.\ embeddings or language models) have prior knowledge about the meaning of the concepts. Some approaches have been studied which only rely on the structure of the ontology itself for learning ontology embeddings, taking inspiration from knowledge graph embedding models \cite{DBLP:conf/ijcai/KulmanovLYH19,DBLP:conf/aaaiss/MondalBM21,DBLP:conf/semweb/XiongPTNS22,DBLP:journals/corr/abs-2202-14018,DBLP:journals/corr/abs-2301-11118}. The special case of embedding taxonomies has also received extensive interest, given the practical importance of taxonomies \cite{vilnis-etal-2018-probabilistic,DBLP:conf/nips/NickelK17,DBLP:conf/icml/GaneaBH18,le-etal-2019-inferring}. Yet another line of work has focused on learning concept representations using word embedding models \cite{mikolov-etal-2013-linguistic}, where the key idea is to view ontology axioms as sentences \cite{DBLP:journals/bioinformatics/SmailiGH18,DBLP:journals/bioinformatics/SmailiGH19,DBLP:journals/ml/ChenHJHAH21}. Finally, when a sufficiently set of factual assertions is available (i.e.\ an ABox), we can also find plausible ontology rules by relying on standard rule learning techniques \cite{DBLP:journals/apin/IannonePF07,DBLP:conf/ilp/FanizzidE08,DBLP:journals/ws/BuhmannLW16,DBLP:conf/aaai/SarkerH19}.

LLMs have not previously been considered for ontology completion, to the best of our knowledge. However, \citet{DBLP:journals/corr/abs-2309-07172} carried out a preliminary study into the potential of LLMs for the related problem of ontology alignment, i.e.\ mapping the concepts from one ontology onto the corresponding concepts from another ontology. They obtained mixed results with Flan-T5-XXL and ChatGPT, with both models failing to consistently outperform a fine-tuned BERT method (in a zero-shot setting). 

\section{NLI Based Models}\label{secNLImodels}
We experiment with three NLI based approaches: (i) the BERTSubs model \cite{DBLP:journals/www/ChenHGJDH23} that was implemented in the DeepOnto library\footnote{https://krr-oxford.github.io/DeepOnto/}; (ii) fine-tuned LLMs; and (iii) ChatGPT and GPT-4 in a zero-shot setting. We now describe these strategies in more detail.

\paragraph{DeepOnto}
BERTSubs solves the following task: given a candidate rule, predict whether it is valid or not. It relies on a fine-tuned LM from the BERT family. The model takes a verbalisation of the rule as input and is fine-tuned as a binary classifier.\footnote{Note that NLI systems are typically ternary classifiers, with entailment, contradiction and neutral as the possible option. In description logic, contradiction is expressed using disjointness rules of the form $X\sqcap Y\sqsubseteq \bot$, which we verbalise as ``$X$ and $Y$ implies contradiction''. In this way, entailment and contradiction is predicted using the same binary classifier.} We rely on the implementation of BERTSubs from the DeepOnto library \cite{DBLP:journals/corr/abs-2307-03067}. For a rule of the form $X\sqsubseteq Y$, we use DeepOnto's verbaliser to obtain a textual description of the concepts $X$ and $Y$. This verbaliser tokenises multi-term expressions and describes the logical structure of complex concepts. For instance, $\mathsf{RedWine}$ becomes ``red wine'' while $\mathsf{Wine} \sqcap \exists \mathsf{hasColor}.\mathsf{Red}$ becomes ``wine that has color red''. To check the validity of $X\sqsubseteq Y$, an input of the form ``[CLS] $d_X$ [SEP] $d_Y$ [SEP]'' is used, with $d_X$ and $d_Y$ the descriptions of $X$ and $Y$ respectively. The prediction is then made using a binary classification head on top of the output representation of the [CLS] token. Two variants are considered which additionally include some of the ontology context of $X$ and $Y$ as part of the input. Since the impact of these variants was found to be limited by \citet{DBLP:journals/www/ChenHGJDH23} we do not consider them in this paper. We use variants with RoBERTa-base and RoBERTa-large \cite{DBLP:journals/corr/abs-1907-11692} as the LM. 

\paragraph{LLMs} 
NLI based methods rely on the LM's internal knowledge to assess the plausibility of a given rule. As such, we can expect that recent LLMs will perform substantially better than BERT-based encoders. While \citet{DBLP:journals/corr/abs-2309-07172} obtained somewhat disappointing results with LLMs, their analysis was limited to the zero-shot setting. Instead, to allow for a more direct comparison, we will use LLMs that are fine-tuned in a similar way to BERTSubs. To fine-tune the LLMs, we use the following prompt:
\begin{myquote}
\textit{Classify the text into True or False. Reply with only one word: True or False. Determine if the following statement is valid: \textsc{[Rule Body]} implies \textsc{[Rule Head]}.}
\end{myquote}
where \textsc{[Rule Body]} and \textsc{[Rule Head]} are the verbalisations of the rule body (i.e.\ left-hand side) and head, obtained with DeepOnto. For instance, for the rule $\mathsf{CheninBlanc} \sqsubseteq \exists \mathsf{hasFlavor}. \mathsf{Moderate}$, the last part of the prompt becomes: \textit{Chenin Blanc implies something that has flavor Moderate}. 


\paragraph{ChatGPT} 
To complement our experiments with fine-tuned LLMs, we also report results for ChatGPT (\texttt{gpt-3.5-turbo}) and GPT-4 (\texttt{gpt-4-turbo}), where we use the same prompt as for the fine-tuned LLMs. 

\section{Concept Embedding Based Models}\label{secConceptEmbeddingBasedModels}
We now recall the approach from  \citet{DBLP:conf/semweb/LiBS19}, which uses a GNN with pre-computed concept embeddings for predicting plausible rules.
We first explain how \emph{rule templates} are used to construct a graph representation of the ontology (Section \ref{secRuleTemplates}). In Section \ref{secGNNModel} we then describe the corresponding GNN models that we will rely on in our experiments. Apart from the choice of the GNN model, the choice of pre-trained concept embeddings also plays a critical role in the overall performance of the model. Section \ref{secConceptEmbeddings} describes the concept embeddings which we have considered in our analysis.

\subsection{Rule Templates}\label{secRuleTemplates}
A unary rule template is obtained by replacing one of the concepts appearing in a rule from the ontology by a placeholder. For instance, the rules \eqref{eqExBritishScientist1}--\eqref{eqExBritishScientist3} are all instances of the following template:
$$
\rho(X) = X \sqcap (\exists \mathsf{livesIn.UK} \sqsubseteq \mathsf{UKScientist})
$$
The notion of rule template allows us to treat the problem of predicting missing rules as a binary classification problem: for a given concept $X$ decide whether $\rho(X)$ is a valid rule or not. A key limitation of this approach is that it can only be used to predict instances of unary templates which are witnessed in the training data. For instance, if \eqref{eqExBritishScientist1}--\eqref{eqExBritishScientist3} are in the training data, we may predict that \eqref{eqRulePhysicistUK} is also plausible. However, consider the following rule:
$$
\mathsf{Biologist} \sqcap (\exists\, \mathsf{livesIn}.\mathsf{France})\sqsubseteq \mathsf{FrenchScientist} 
$$
We cannot predict this rule in the same way, because it differs from the known rules in two places.  \citet{DBLP:conf/semweb/LiBS19} therefore also considered binary templates, which replace two concepts in a rule with a placeholder. For instance, we can consider:
$$
\rho(X,Y) = \mathsf{Biologist} \sqcap (\exists \mathsf{livesIn}.X) \sqsubseteq Y
$$
The rule \eqref{eqExBritishScientist1} is an instance of this template with $X=\mathsf{UK}$ and $Y=\mathsf{UKScientist}$. 
Each binary template defines a classification problem on concept pairs $(X,Y$), i.e.\ decide whether $\rho(X,Y)$ is valid. 

One issue with the binary templates is that many of the rules in a typical ontology are basic subsumptions of the form $A\sqsubseteq B$, which give us the trivial binary template $\rho(X,Y) = X\sqsubseteq Y$. Following \citet{DBLP:conf/semweb/LiBS19}, we will therefore rely on \emph{typed} binary templates. When typed templates are used, instead of replacing a concept $A$ by a placeholder $X$, we replace it by the conjunction $X\sqcap A'$ where $A'$ is a direct parent of $A$ (i.e.\ $A'$ is such that we have the rule $A\sqsubseteq A'$ in the ontology). A basic subsumption $A\sqsubseteq B$ then leads to a binary template of the form $\rho(X,Y) = X\sqcap A' \sqsubseteq Y \sqcap B'$, where $A'$ and $B'$ are direct parents of $A$ and $B$. If $A$ and $B$ have multiple direct parents, then we consider each of the corresponding typed templates.

\subsection{GNN Model}\label{secGNNModel}
When using rule templates, the problem of predicting plausible rules reduces to that of classifying concepts or concept pairs. While we could, in principle, directly use pre-trained concept embeddings for this purpose, better results can be achieved by contextualising these embeddings using a GNN. Specifically, we consider a graph with one node for each concept appearing in the training set. Two concepts are connected by an edge if they appear in the same rule. \citet{DBLP:conf/semweb/LiBS19} used edges of different types, corresponding to binary templates. In other words, their graph structure reflects which kinds of rules two concepts co-occur in. While their graph is more informative, learning with multi-relational graphs is harder, especially considering that the amount of training data that we have available is typically limited. For this reason, we consider a simple graph, only capturing whether two concepts appear in the same rule or not. As we will see, this leads to similar empirical results, while it makes the model more efficient. 

We consider two variants of the model, depending on whether unary or binary templates are used. Let us first consider the unary template model. Let $\mathcal{U}$ be the set of unary templates which are witnessed in the given ontology. For each of these unary templates $\rho$, we train a binary classifier to categorise concepts into positive examples (i.e.\ those concepts $X$ for which $\rho(X)$ is believed to be a valid rule) and negative examples.
Specifically, for each concept $X$ and template $\rho\in\mathcal{U}$, we estimate the probability that $\rho(X)$ is a valid rule as follows
$$
\textit{conf}(\rho,X) = \sigma(\mathbf{x}\cdot \mathbf{a}_{\rho} + b_{\rho})
$$
where $\mathbf{x}\in\mathbb{R}^n$ is the final-layer embedding of concept $X$ in the GNN, $\mathbf{a}_{\rho}\in \mathbb{R}^n$ and $b_{\rho}\in\mathbb{R}$, and $\sigma$ denotes the sigmoid activation function. To classify a given rule $r$ as valid or not, we first determine all templates $\rho$ and concepts $X$ for which $r=\rho(X)$. Let us write $\rho_1(X_1),...,\rho_m(X_m)$ for these template-concept combinations. The probability that $r$ is a valid rule is then estimated as:
$$
p(r) = \max_{i=1}^m \textit{conf}(\rho_i,X_i)
$$
Note that if $m=0$, i.e.\ $r$ is not an instance of any of the unary templates, then $p(r)=0$. Different from \citet{DBLP:conf/semweb/LiBS19}, we train the model using binary cross-entropy at the level of rule predictions:
$$
\mathcal{L} = - \sum_{r} y_r \log p(r) + (1-y_r)\log (1-p(r))
$$
where the summation ranges over the rules $r$ in the training set (see Section \ref{secDataset}), and we define $y_r=1$ if $r$ is a positive example and $y_r=0$ otherwise. 

For the binary template model, we need to predict whether $\rho(X,Y)$ is a valid rule. To this end, we use the scoring function from the distmult \cite{DBLP:journals/corr/YangYHGD14a} knowledge graph embedding model.\footnote{We also experimented with the TransE \cite{DBLP:conf/nips/BordesUGWY13}scoring function as an alternative and found the results to be broadly similar. An analysis can be found in the appendix.}
In particular, we have:
%
$$
\textit{conf}(\rho,X,Y) = \sigma(\mathbf{x}^{T} \mathbf{M}_{\rho} \mathbf{y})
$$
where $\mathbf{M}_{\rho} \in \mathbb{R}^{n \times n}$ is a diagonal matrix, and $\mathbf{x},\mathbf{y}\in\mathbb{R}^n$ are the final-layer embeddings of concepts $X$ and $Y$ in the GNN. The binary template model is also trained using binary cross-entropy.

\subsection{Concept Embeddings}\label{secConceptEmbeddings}
We experiment with several types of pre-trained concept embeddings. First, we consider standard word embeddings models: Skip-gram \cite{mikolov-etal-2013-linguistic}, GloVe \cite{pennington-etal-2014-glove} 
and Numberbatch \cite{DBLP:conf/aaai/SpeerCH17}\footnote{Glove trained on Common Crawl (\url{https://nlp.stanford.edu/projects/glove/}), Skip-gram trained on Google News (\url{https://code.google.com/archive/p/word2vec/}), and Numberbatch  from \url{https://conceptnet.s3.amazonaws.com/downloads/2019/numberbatch/numberbatch-en-19.08.txt.gz}}. Second, we consider methods which rely on fine-tuned LM encoders to map the name of a concept to their embedding: MirrorBERT\footnote{\url{https://huggingface.co/cambridgeltl/mirror-bert-base-uncased-word}} \cite{liu-etal-2021-fast}, which was trained in a self-supervised fashion, and the bi-encoder model from \citet{gajbhiye-etal-2022-modelling}, which we will refer to as BiEnc\footnote{\url{https: //github.com/amitgajbhiye/biencoder_ concept_property}}. Finally, we use two methods which obtain embeddings by finding mentions of the concept name in Wikipedia and aggregating their contextualised representation: MirrorWiC\footnote{\url{https://huggingface.co/cambridgeltl/mirrorwic-bert-base-uncased}} \cite{liu-etal-2021-mirrorwic}, which is self-supervised, and ConCN\footnote{\url{https://github.com/lina-luck/semantic_concept_embeddings} }\cite{DBLP:conf/sigir/LiKBS23}, which was trained using distant supervision from ConceptNet.

\section{Dataset}\label{secDataset}
Existing approaches for evaluating ontology completion methods have some important limitations. First, the GNN based method from \citet{DBLP:conf/semweb/LiBS19} is evaluated based on the accuracy of the classifiers associated with the unary and binary templates, which does not allow us to analyse the effectiveness of using rule templates in itself. Second, benchmarks such as ONTOLAMA consider individual rules in isolation, which means that they cannot be used to evaluate methods that need a given ontology. Third, existing methods are usually evaluated as classification problems, where positive examples are rules that were held out from a given ontology and negative examples are generated by randomly corrupting positive examples. The use of randomly generated negative examples has important drawbacks. For instance, they are often relatively easy to identify, especially if they were obtained by swapping concepts for semantically distinct alternatives. As another limitation, some of the randomly negative rules might actually correspond to semantically valid rules. To address these limitations, we have created a new benchmark with manually annotated negative examples.

\begin{table}[]
\centering
\setlength\tabcolsep{3pt}
\fontsize{8.5pt}{9.8pt}\selectfont
\label{data statistics}
\begin{tabular}{l ccccccc}
\toprule
          & \multicolumn{2}{c}{\textbf{Training}} & \multicolumn{2}{c}{\textbf{Dev}} & \multicolumn{2}{c}{\textbf{Test}} & \textbf{IAA}\\
           \cmidrule(lr){2-3}\cmidrule(lr){4-5} \cmidrule(lr){6-7} \cmidrule(lr){8-8} 
          & \textbf{pos} & \textbf{neg} & \textbf{pos} & \textbf{neg}  & \textbf{pos} & \textbf{neg}  & $\boldsymbol\kappa$\\ 
\midrule
Wine      & 69         & 319       & 10 & 31   & 18        & 15    &  80.0    \\ 
Economy   & 384        & 1744      & 44 & 174    & 96       & 81    &  82.0    \\ 
Olympics  & 135        & 621       & 14 & 63    & 34       & 29    &  83.0    \\ 
Transport & 615        & 2416      & 135 & 142    & 154      & 145   &  81.0    \\ 
SUMO      & 4377       & 21624     & 735 & 749    & 1095     & 998   &  63.0    \\ 
FoodOn    & 45013      & 221429    & 2260 & 2190    & 2370     & 2155  &  62.0    \\ 
GO        & 103184     & 494708    & 5326 & 5012    & 5431     & 5044  &  58.0    \\ 
\bottomrule  
\end{tabular}
\caption{Overview of the dataset, showing the number of positive and negative examples in the training and test split, as well as the inter-annotator agreement (IAA), in terms of Cohen's $\kappa$, for the negative test examples.\label{tabDatasetStatistics}}
\end{table}

\paragraph{Ontologies} The benchmark is based on seven well-known ontologies. In particular, we have included the five ontologies which were used by \citet{DBLP:conf/semweb/LiBS19}: Wine\footnote{\url{https://www.w3.org/TR/2003/PR-owl-guide-20031215/wine}}, Economy\footnote{\url{http://reliant.teknowledge.com/DAML/Economy.owl}}, Olympics\footnote{\url{https://swat.cse.lehigh.edu/resources/onto/olympics.owl}}, Transport\footnote{\url{http://reliant.teknowledge.com/DAML/Transportation.owl}} and SUMO\footnote{\url{SUMO  https://www.ontologyportal.org/}}. In addition, we have included two ontologies that were used by \citet{DBLP:journals/www/ChenHGJDH23}: FoodOn\footnote{\url{https://obofoundry.org/ontology/foodon.html}} and the Gene Ontology (GO)\footnote{\url{http://purl.obolibrary.org/obo/go.owl}}. These ontologies cover a number of different settings. For instance, Wine, Economy, Olympics and Transport are small domain-specific ontologies. SUMO is used as a representative example of a larger general-domain ontology. Finally, FoodOn and GO are considerably larger than all the others, while being focused on a specialised domain. For Wine, Economy, Olympics, Transport and SUMO, we keep 20\%  of the rules for testing. In the case of FoodOn and GO, we keep 5\% for testing. The remaining rules are split into training and development sets. Some basic statistics of the considered ontologies are summarised in Table \ref{tabDatasetStatistics}.





\paragraph{Negative Training Examples}
For the training set, using manually annotated negative examples is not feasible, due to the large number of rules which would have to be checked. Therefore, we instead rely on  corrupting rules from the given ontology, using similar strategies as in previous work \cite{DBLP:conf/semweb/LiBS19,DBLP:journals/www/ChenHGJDH23}. In the following, we will use notations such as $\alpha$ and $\beta$ to denote arbitrary rule bodies and heads, and notations such as $C$ and $D$ to denote concept names. We use the following strategies.
    (i) For each rule of the form $C\sqsubseteq D$ in the ontology, we add $D\sqsubseteq C$ as a negative rule. 
    (ii) For each rule of the form $\alpha_1 \sqsubseteq \beta_1$, we randomly select another rule from the ontology of the form $\alpha_2 \sqsubseteq \beta_2$ and we generate the corrupted rules $\alpha_1 \sqsubseteq \beta_2$ and $\alpha_2\sqsubseteq \beta_1$.  
    (iii) For each rule of the form $C \sqsubseteq D$, we randomly replace $C$ or $D$ by another concept, which is randomly sampled from all concepts appearing in the ontology.
    (iv) For each rule of the form $C\sqsubseteq D$ we generate the constraint $C\sqcap D \sqsubseteq \bot$ (encoding that $C$ and $D$ are disjoint).
In all cases, when we corrupt rules using these strategies, we only add these rules as negative examples to the training set if they are not entailed by the positive rules from the training split. 

\paragraph{Negative Test Examples}
To obtain negative examples for the test set, we rely on human annotators to ensure that the corrupted rules are indeed negative examples. Moreover, we aim to select hard negative examples, given that random corruption often leads to nonsensical rules which are too easy to identify. Specifically, for each positive rule $\alpha \sqsubseteq \beta$, we randomly select one of the concepts $C$ appearing in that rule and replace it with another concept $D$. Rather than selecting this concept arbitrarily, we choose $D$ to be among the 5 most similar concepts to $C$, in terms of the cosine similarity between the fastText embeddings of the corresponding names\footnote{Specifically, we used the embeddings that were trained on Common Crawl, which are available from \url{https://fasttext.cc/docs/en/english-vectors.html}.}. 
Note that selecting similar concepts increases the chances that the corrupted rule is actually a valid rule, which means that human annotation is critical in this case. Each corrupted rule was checked by two annotators, who were trained in formal knowledge representation and were fluent in English. The agreement between the annotators is reported in Table \ref{tabDatasetStatistics}. We only keep negative examples that were annotated as being genuine negatives by both of the annotators.


\begin{table*}[t]
\centering
\footnotesize
\setlength\tabcolsep{5pt}
\begin{tabular}{lcccccccc}
\toprule
& \textbf{Wine} & \textbf{Economy} & \textbf{Olympics} & \textbf{Transport} & \textbf{SUMO} & \textbf{FoodOn} & \textbf{GO} & \textbf{Average} \\
\midrule
\multicolumn{9}{c}{\textsc{NLI Based Models}}\\
\midrule
RoBERTa-base  & 57.8 & 78.5 & 76.9 & 65.6 & 76.8 & 76.2 & 72.9 & 72.1\\
RoBERTa-large & 76.5 & 79.4 & 79.3 & 75.6 & 77.3 & 78.3 & 74.6 & 77.3\\
\midrule
Llama2-7B  & 45.0 & 63.3 & 55.1 & 60.0 & 64.0 & 75.1 & 72.7 & 62.2\\
Llama2-13B  & 52.2 & 68.2 & 55.6 & 62.0 & 69.5 & 78.2 & 77.2 & 66.1\\
Mistral-7B  & 62.3 & 71.9 & 66.5 & 70.3 & 72.5 & 77.5 & 75.1 & 70.9\\
Vicuna-13B & 54.1 & 78.4 & 73.0 & 69.4 & 72.5 & 77.3 & 76.8 & 71.6\\
\midrule
ChatGPT & 50.8 & 66.5 & 69.5 & 56.1 & 65.6 & 60.3 & 61.7 & 61.5\\
GPT-4 & 63.7 & 74.8 & 81.0 & 62.4 & 74.2 & 75.7 & 76.7 & 72.6\\
\midrule
\multicolumn{9}{c}{\textsc{Concept Embedding Based Models}}\\
\midrule
R-GCN (UT) & 84.8 & 71.6 & 64.7 & 74.9 & 69.4 & 70.8 & 71.3 & 72.5\\
R-GCN (BT) & 79.9 & 16.5 & 28.5 & 54.0 & 3.7 & 25.7 & 28.3 & 33.8\\
GCN (UT) & 84.8 & 73.4 & 68.9 & 76.9 & 67.2 & 71.5 & 71.9 & 73.5\\
GCN (BT) & 78.9 & 16.6 & 28.5 & 52.8 & 3.7 & 26.3 & 28.7 & 33.6\\
GAT (UT) & 84.8 & 68.0 & 68.6 & 74.3 & 69.3 & 69.4 & 70.6 & 72.1\\
GAT (BT) & 87.1 & 16.6 & 29.2 & 55.6 & 3.7 & 25.3 & 27.5 & 35.0\\
GATv2 (UT) & \textbf{88.2} & 66.9 & 65.5 & 74.7 & 69.5 & 69.7 & 70.9 & 72.2\\
GATv2 (BT) & 86.4 & 16.5 & 29.2 & 55.7 & 3.7 & 25.8 & 26.7 & 34.9\\
\midrule
\multicolumn{9}{c}{\textsc{Hybrid Models}}\\
\midrule
GCN (UT + BT)  & 84.8 & 73.7 & 67.1 & 77.2 & 67.6 & 71.5 & 72.4 & 73.5\\
GCN (UT) + Llama2-13B  & 84.9 & 80.5 & 81.2 & 78.3 & 79.4 & 80.1 & 77.6 & 80.3\\
GCN (BT) + Llama2-13B  & 82.4 & 54.9 & 56.4 & 63.9 & 20.2 & 41.1 & 58.2 & 53.9\\
GCN (UT + BT) + Llama2-13B  & 85.1 & \textbf{80.8} & \textbf{81.6} & \textbf{78.9} & \textbf{80.2} & \textbf{80.9} & \textbf{79.2} & \textbf{81.0}\\
\bottomrule
\end{tabular}
\caption{Overview of the main results in terms of F1 ($\%$). For the GNN models, ConCN embeddings were used as input features. The NLI based models are trained on each ontology separately. \label{tabMainResults}}
\end{table*}

\begin{table*}[t]
\centering
\footnotesize
\begin{tabular}{llcccccccc}
\toprule
&& \textbf{Wine} & \textbf{Economy} & \textbf{Olympics} & \textbf{Transport} & \textbf{SUMO} & \textbf{FoodOn} & \textbf{GO} & \textbf{Average} \\
\midrule
\parbox[t]{2mm}{\multirow{7}{*}{\rotatebox[origin=c]{90}{\textbf{Unary templates}}}}
&Skip-gram & 50.3 & 52.2 & 47.7 & 48.3 & 51.6 & 56.4 & 55.1 & 51.7\\
&GloVe  & 51.2 & 53.4 & 48.9 & 50.2 & 53.7 & 58.7 & 57.6 & 53.4\\
&Numberbatch  & 82.8 &72.3 & 67.5 &75.2  &66.9 & 70.8 & 70.5 & 72.3\\
&MirrorBERT  & 81.3 &71.5 & 65.4 &70.3  &63.4 & 68.1 & 67.1 & 69.6\\
&MirrorWiC  & 82.4 &71.9 & 66.2 & 71.6  & 64.5 & 68.8 & 68.3 & 70.5\\
&BiEnc  & 83.2 &72.9 & 68.2 & 75.4  & 66.8 & 71.2 & 70.9 & 72.7\\
&ConCN & 84.8 & 73.4 & 68.9 & 76.9 & 67.2 & 71.5 & 71.9 & 73.5\\
\midrule
\parbox[t]{2mm}{\multirow{7}{*}{\rotatebox[origin=c]{90}{\textbf{Binary templates}}}}
&Skip-gram & 47.8 & 13.1 & 19.9 & 42.8 & 2.6 & 18.7 & 18.3 & 23.3\\
&GloVe & 49.1 & 13.2 & 20.4 & 43.2 & 2.6 & 19.2 & 18.9 & 23.8\\
&Numberbatch & 77.3 & 16.1 & 27.9 & 51.6 & 3.7 & 25.9 & 27.9 & 32.9\\
&MirrorBERT & 74.5 & 15.7 & 26.4 & 50.3 & 3.6 & 23.7 & 24.8 & 31.3\\
&MirrorWiC & 75.2 & 15.9 & 26.5 & 50.9 & 3.6 & 24.5 & 25.4 & 31.7\\
&BiEnc & 76.3 & 16.2 & 28.1 & 52.6 & 3.7 & 25.8 & 28.1 & 33.0\\
&ConCN & 78.9 & 16.6 & 28.5 & 52.8 & 3.7 & 26.3 & 28.7 & 33.6\\
\bottomrule
\end{tabular}
\caption{Analysis of different pre-trained concept embeddings. All results are obtained with the GCN model. \label{secConceptEmbeddingsAnalysis}}
\end{table*}


\section{Experiments}
We now experimentally compare the models from Sections \ref{secNLImodels} and \ref{secConceptEmbeddingBasedModels}.\footnote{Our dataset and implementation will be shared upon acceptance.}

\paragraph{Models}
We experiment with the NLI based models from the DeepOnto library, which correspond to fine-tuned \textbf{RoBERTa-base} and \textbf{RoBERTa-large} encoders. Regarding the fine-tuned LLMs, we experiment with the 7B and 13B parameter \textbf{Llama 2}\footnote{\url{https://llama.meta.com}} models, as well as the 7B parameter \textbf{Mistral}\footnote{\url{https://github.com/mistralai/mistral-src}} and 13B parameter \textbf{Vicuna}\footnote{\url{https://lmsys.org/blog/2023-03-30-vicuna/}} models.  We use the base versions of these models, rather than the instruction fine-tuned variants, as we found the former to perform somewhat better.\footnote{An analysis of the performance of such model variants can be found in the appendix.}
For the concept embedding based approach we consider three standard GNN architectures: \textbf{GCN} \cite{DBLP:conf/iclr/KipfW17}, \textbf{GAT} \cite{DBLP:conf/iclr/VelickovicCCRLB18} and \textbf{GATv2} \cite{DBLP:conf/iclr/Brody0Y22}. We also compare with the model from \citet{DBLP:conf/semweb/LiBS19}, which uses a graph with different edge types, corresponding to the binary templates, and relies on an \textbf{R-GCN} \cite{DBLP:conf/esws/SchlichtkrullKB18} to take these edge types into account. Unless specified otherwise, we use the ConCN concept embeddings \cite{DBLP:conf/sigir/LiKBS23} as input features for the GNNs.

\paragraph{Results}
The main results are summarised in Table \ref{tabMainResults}. For this experiment, the NLI models have been fine-tuned on each ontology separately.\footnote{The other possibility is to train a single model on the joint training sets of all ontologies. An analysis of this variant can be found in the appendix.}  Somewhat surprisingly, perhaps, the RoBERTa-large model from DeepOnto outperforms the other NLI based models, as well as the concept embedding based models.  Among the fine-tuned LLMs, Vicuna-13B achieves the best results. For the concept embedding based models, the unary templates generally perform much better than the binary templates. The different GNN architectures perform similarly, with the best overall results achieved by the GCN. 
In particular, the GCN slightly outperforms the less efficient R-GCN approach from \citet{DBLP:conf/semweb/LiBS19}. 

\paragraph{Hybrid Methods}
The GNN models can only predict a given rule if it is an instance of a rule template that is witnessed in the training data. As a result of this limitation, there are a significant number of rules from the test set that can never be predicted. At the same time, the LLM based methods are limited because they are harder to adapt to a given ontology. Since the GNN and LLM models have complementary strengths, it is natural to consider the following hybrid strategy for predicting rules: if $r$ is the instance of some rule template, then we predict the validity of $r$ using the GNN model. Otherwise, we predict the validity of $r$ using a different model. Table \ref{tabMainResults} shows the performance of this hybrid strategy. For the GCN (UT + BT) variant, we use GCN (UT) as the main model and GCN (BT) as the fallback model. For GCN (UT + BT) + Llama2-13B, we use Llama2-13B as a second fall-back model, in case there are no unary templates nor any binary templates that can be used. As can be seen, this hybrid approach is highly effective, with GCN (UT + BT) + Llama2-13B overall achieving the best results. This clearly shows the complementary nature of the NLI and concept embedding based approaches.

\paragraph{Comparison of Concept Embeddings}
For the GNN models, Table \ref{tabMainResults} only reports results for the ConCN concept embeddings. To analyse the impact of this choice, 
Table \ref{secConceptEmbeddingsAnalysis} compares the results for different concept embedding choices. For this analysis, we have used the GCN model. As can be seen, the results are highly sensitive to the quality of the concept embeddings, with traditional word embeddings models such as Skip-gram and GloVe leading to considerably weaker results.

\paragraph{Qualitative Analysis}
When inspecting examples of rules\footnote{A list of such rules is shown in the appendix.} that were correctly predicted by the GCN (with unary templates and ConCN embeddings) but not by Llama2-13B, we see several cases involving domain-specific concepts, e.g.\ $\textit{Rings implies Artistic Gymnastics}$, which only make sense within the context of the given ontology (i.e.\ Olympics). We also see cases involving $\exists$, which often sound less natural when verbalised, e.g.\ \textit{Beaujolais implies something that has sugar Dry}. Conversely, looking at examples that Llama2-13B correctly predicted, but not the GCN, we see many examples that are almost tautological, e.g.\ \textit{Sauternes implies something located in Sauterne Region}. Such rules are easy to identify by NLI models, but GNN models can fail on such cases if they lack the required template. NLI models also do well on examples that benefit from the general background knowledge captured by LLMs, e.g.\ \textit{Fire Boat implies Emergency Vehicle}.

\section{Conclusions}
We have considered the problem of finding plausible rules which are missing from a given ontology, as a generalisation of the widely-studied problem of taxonomy expansion. While ontology completion has already been studied in previous work, different families of methods have been studied more or less in isolation, with no previous empirical comparison between them. Moreover, the benchmarks that have thus far been used involve distinguishing valid rules from randomly corrupted rules, which has several drawbacks (e.g.\ randomly corrupted rules are often easier to detect). To address these issues, we have introduced a new benchmark with hard negatives, which were manually verified by human annotators. We then compared the two main families of ontology completion methods: NLI based methods and GNN based methods. Beyond existing NLI based methods, we also presented the first analysis of LLMs for ontology completion. 
Finally, we found hybrid strategies, which have not previously been considered, to achieve the best results, using a GNN based method for those cases where it can be used and falling back to an LLM otherwise. 

\section*{Limitations}
The area of ontology completion is considerably less mature than related areas such as taxonomy expansion and knowledge graph completion. As such, the methods we have analysed in this paper should be seen as baselines for future work. For instance, we expect that much better hybrid strategies can be developed, which combine the knowledge captured by LLMs with models that take into account the structure of the ontology. The evaluation results of the LLM models themselves should also be seen as lower bounds. For instance, while we have attempted to construct reasonable prompts, it is likely that better prompting strategies can be found.

In this paper, we have treated ontology completion as a binary classification problem, deciding whether a given candidate rule is valid or not. However, in practice, we also need a mechanism for generating suitable candidate rules. The template based approach can be used in a straightforward way for this purpose. While it is likely that LLMs can also be successfully leveraged for generating rules, rather than classifying them, studying how this can best be done is left for future work.

\bibliography{custom,anthology}

\begin{thebibliography}{53}
\expandafter\ifx\csname natexlab\endcsname\relax\def\natexlab#1{#1}\fi

\bibitem[{Baader et~al.(2009)Baader, Horrocks, and Sattler}]{DBLP:series/ihis/BaaderHS09}
Franz Baader, Ian Horrocks, and Ulrike Sattler. 2009.
\newblock \href {https://doi.org/10.1007/978-3-540-92673-3\_1} {Description logics}.
\newblock In Steffen Staab and Rudi Studer, editors, \emph{Handbook on Ontologies}, International Handbooks on Information Systems, pages 21--43. Springer.

\bibitem[{Beltagy et~al.(2016)Beltagy, Roller, Cheng, Erk, and Mooney}]{beltagy-etal-2016-representing}
I.~Beltagy, Stephen Roller, Pengxiang Cheng, Katrin Erk, and Raymond~J. Mooney. 2016.
\newblock \href {https://doi.org/10.1162/COLI_a_00266} {Representing meaning with a combination of logical and distributional models}.
\newblock \emph{Computational Linguistics}, 42(4):763--808.

\bibitem[{Bordea et~al.(2016)Bordea, Lefever, and Buitelaar}]{bordea-etal-2016-semeval}
Georgeta Bordea, Els Lefever, and Paul Buitelaar. 2016.
\newblock \href {https://doi.org/10.18653/v1/S16-1168} {{S}em{E}val-2016 task 13: Taxonomy extraction evaluation ({TE}x{E}val-2)}.
\newblock In \emph{Proceedings of the 10th International Workshop on Semantic Evaluation ({S}em{E}val-2016)}, pages 1081--1091, San Diego, California. Association for Computational Linguistics.

\bibitem[{Bordes et~al.(2013)Bordes, Usunier, Garc{\'{\i}}a{-}Dur{\'{a}}n, Weston, and Yakhnenko}]{DBLP:conf/nips/BordesUGWY13}
Antoine Bordes, Nicolas Usunier, Alberto Garc{\'{\i}}a{-}Dur{\'{a}}n, Jason Weston, and Oksana Yakhnenko. 2013.
\newblock \href {https://proceedings.neurips.cc/paper/2013/hash/1cecc7a77928ca8133fa24680a88d2f9-Abstract.html} {Translating embeddings for modeling multi-relational data}.
\newblock In \emph{Advances in Neural Information Processing Systems 26: 27th Annual Conference on Neural Information Processing Systems 2013. Proceedings of a meeting held December 5-8, 2013, Lake Tahoe, Nevada, United States}, pages 2787--2795.

\bibitem[{Bouraoui and Schockaert(2019)}]{DBLP:conf/aaai/BouraouiS19}
Zied Bouraoui and Steven Schockaert. 2019.
\newblock \href {https://doi.org/10.1609/AAAI.V33I01.33016228} {Automated rule base completion as bayesian concept induction}.
\newblock In \emph{The Thirty-Third {AAAI} Conference on Artificial Intelligence, {AAAI} 2019, The Thirty-First Innovative Applications of Artificial Intelligence Conference, {IAAI} 2019, The Ninth {AAAI} Symposium on Educational Advances in Artificial Intelligence, {EAAI} 2019, Honolulu, Hawaii, USA, January 27 - February 1, 2019}, pages 6228--6235. {AAAI} Press.

\bibitem[{Brody et~al.(2022)Brody, Alon, and Yahav}]{DBLP:conf/iclr/Brody0Y22}
Shaked Brody, Uri Alon, and Eran Yahav. 2022.
\newblock \href {https://openreview.net/forum?id=F72ximsx7C1} {How attentive are graph attention networks?}
\newblock In \emph{The Tenth International Conference on Learning Representations, {ICLR} 2022, Virtual Event, April 25-29, 2022}. OpenReview.net.

\bibitem[{B{\"{u}}hmann et~al.(2016)B{\"{u}}hmann, Lehmann, and Westphal}]{DBLP:journals/ws/BuhmannLW16}
Lorenz B{\"{u}}hmann, Jens Lehmann, and Patrick Westphal. 2016.
\newblock \href {https://doi.org/10.1016/j.websem.2016.06.001} {Dl-learner - {A} framework for inductive learning on the semantic web}.
\newblock \emph{J. Web Semant.}, 39:15--24.

\bibitem[{Chen et~al.(2021{\natexlab{a}})Chen, Lin, and Klein}]{chen-etal-2021-constructing}
Catherine Chen, Kevin Lin, and Dan Klein. 2021{\natexlab{a}}.
\newblock \href {https://doi.org/10.18653/v1/2021.naacl-main.373} {Constructing taxonomies from pretrained language models}.
\newblock In \emph{Proceedings of the 2021 Conference of the North American Chapter of the Association for Computational Linguistics: Human Language Technologies}, pages 4687--4700, Online. Association for Computational Linguistics.

\bibitem[{Chen et~al.(2023)Chen, He, Geng, Jim{\'{e}}nez{-}Ruiz, Dong, and Horrocks}]{DBLP:journals/www/ChenHGJDH23}
Jiaoyan Chen, Yuan He, Yuxia Geng, Ernesto Jim{\'{e}}nez{-}Ruiz, Hang Dong, and Ian Horrocks. 2023.
\newblock \href {https://doi.org/10.1007/S11280-023-01169-9} {Contextual semantic embeddings for ontology subsumption prediction}.
\newblock \emph{World Wide Web {(WWW)}}, 26(5):2569--2591.

\bibitem[{Chen et~al.(2022)Chen, He, Jim{\'{e}}nez{-}Ruiz, Dong, and Horrocks}]{DBLP:journals/corr/abs-2202-09791}
Jiaoyan Chen, Yuan He, Ernesto Jim{\'{e}}nez{-}Ruiz, Hang Dong, and Ian Horrocks. 2022.
\newblock \href {http://arxiv.org/abs/2202.09791} {Contextual semantic embeddings for ontology subsumption prediction}.
\newblock \emph{CoRR}, abs/2202.09791.

\bibitem[{Chen et~al.(2021{\natexlab{b}})Chen, Hu, Jim{\'{e}}nez{-}Ruiz, Holter, Antonyrajah, and Horrocks}]{DBLP:journals/ml/ChenHJHAH21}
Jiaoyan Chen, Pan Hu, Ernesto Jim{\'{e}}nez{-}Ruiz, Ole~Magnus Holter, Denvar Antonyrajah, and Ian Horrocks. 2021{\natexlab{b}}.
\newblock \href {https://doi.org/10.1007/s10994-021-05997-6} {Owl2vec*: embedding of {OWL} ontologies}.
\newblock \emph{Mach. Learn.}, 110(7):1813--1845.

\bibitem[{d'Amato et~al.(2012)d'Amato, Fanizzi, Fazzinga, Gottlob, and Lukasiewicz}]{DBLP:journals/amai/dAmatoFFGL12}
Claudia d'Amato, Nicola Fanizzi, Bettina Fazzinga, Georg Gottlob, and Thomas Lukasiewicz. 2012.
\newblock \href {https://doi.org/10.1007/s10472-012-9309-7} {Ontology-based semantic search on the web and its combination with the power of inductive reasoning}.
\newblock \emph{Ann. Math. Artif. Intell.}, 65(2-3):83--121.

\bibitem[{Devlin et~al.(2019)Devlin, Chang, Lee, and Toutanova}]{devlin-etal-2019-bert}
Jacob Devlin, Ming-Wei Chang, Kenton Lee, and Kristina Toutanova. 2019.
\newblock \href {https://doi.org/10.18653/v1/N19-1423} {{BERT}: Pre-training of deep bidirectional transformers for language understanding}.
\newblock In \emph{Proceedings of the 2019 Conference of the North {A}merican Chapter of the Association for Computational Linguistics: Human Language Technologies, Volume 1 (Long and Short Papers)}, pages 4171--4186, Minneapolis, Minnesota. Association for Computational Linguistics.

\bibitem[{Fanizzi et~al.(2008)Fanizzi, d'Amato, and Esposito}]{DBLP:conf/ilp/FanizzidE08}
Nicola Fanizzi, Claudia d'Amato, and Floriana Esposito. 2008.
\newblock \href {https://doi.org/10.1007/978-3-540-85928-4\_12} {{DL-FOIL} concept learning in description logics}.
\newblock In \emph{Inductive Logic Programming, 18th International Conference, {ILP} 2008, Prague, Czech Republic, September 10-12, 2008, Proceedings}, volume 5194 of \emph{Lecture Notes in Computer Science}, pages 107--121. Springer.

\bibitem[{Fu et~al.(2014)Fu, Guo, Qin, Che, Wang, and Liu}]{fu-etal-2014-learning}
Ruiji Fu, Jiang Guo, Bing Qin, Wanxiang Che, Haifeng Wang, and Ting Liu. 2014.
\newblock \href {https://doi.org/10.3115/v1/P14-1113} {Learning semantic hierarchies via word embeddings}.
\newblock In \emph{Proceedings of the 52nd Annual Meeting of the Association for Computational Linguistics (Volume 1: Long Papers)}, pages 1199--1209, Baltimore, Maryland. Association for Computational Linguistics.

\bibitem[{Gajbhiye et~al.(2022)Gajbhiye, Espinosa-Anke, and Schockaert}]{gajbhiye-etal-2022-modelling}
Amit Gajbhiye, Luis Espinosa-Anke, and Steven Schockaert. 2022.
\newblock \href {https://aclanthology.org/2022.coling-1.349} {Modelling commonsense properties using pre-trained bi-encoders}.
\newblock In \emph{Proceedings of the 29th International Conference on Computational Linguistics}, pages 3971--3983, Gyeongju, Republic of Korea. International Committee on Computational Linguistics.

\bibitem[{Ganea et~al.(2018)Ganea, B{\'{e}}cigneul, and Hofmann}]{DBLP:conf/icml/GaneaBH18}
Octavian{-}Eugen Ganea, Gary B{\'{e}}cigneul, and Thomas Hofmann. 2018.
\newblock \href {http://proceedings.mlr.press/v80/ganea18a.html} {Hyperbolic entailment cones for learning hierarchical embeddings}.
\newblock In \emph{Proceedings of the 35th International Conference on Machine Learning, {ICML} 2018, Stockholmsm{\"{a}}ssan, Stockholm, Sweden, July 10-15, 2018}, volume~80 of \emph{Proceedings of Machine Learning Research}, pages 1632--1641. {PMLR}.

\bibitem[{He et~al.(2023{\natexlab{a}})He, Chen, Dong, and Horrocks}]{DBLP:journals/corr/abs-2309-07172}
Yuan He, Jiaoyan Chen, Hang Dong, and Ian Horrocks. 2023{\natexlab{a}}.
\newblock \href {https://doi.org/10.48550/ARXIV.2309.07172} {Exploring large language models for ontology alignment}.
\newblock \emph{CoRR}, abs/2309.07172.

\bibitem[{He et~al.(2023{\natexlab{b}})He, Chen, Dong, Horrocks, Allocca, Kim, and Sapkota}]{DBLP:journals/corr/abs-2307-03067}
Yuan He, Jiaoyan Chen, Hang Dong, Ian Horrocks, Carlo Allocca, Taehun Kim, and Brahmananda Sapkota. 2023{\natexlab{b}}.
\newblock \href {https://doi.org/10.48550/ARXIV.2307.03067} {Deeponto: {A} python package for ontology engineering with deep learning}.
\newblock \emph{CoRR}, abs/2307.03067.

\bibitem[{He et~al.(2023{\natexlab{c}})He, Chen, Jim{\'{e}}nez{-}Ruiz, Dong, and Horrocks}]{DBLP:journals/corr/abs-2302-06761}
Yuan He, Jiaoyan Chen, Ernesto Jim{\'{e}}nez{-}Ruiz, Hang Dong, and Ian Horrocks. 2023{\natexlab{c}}.
\newblock \href {https://doi.org/10.48550/arXiv.2302.06761} {Language model analysis for ontology subsumption inference}.
\newblock \emph{CoRR}, abs/2302.06761.

\bibitem[{Hearst(1992)}]{hearst-1992-automatic}
Marti~A. Hearst. 1992.
\newblock \href {https://aclanthology.org/C92-2082} {Automatic acquisition of hyponyms from large text corpora}.
\newblock In \emph{{COLING} 1992 Volume 2: The 14th {I}nternational {C}onference on {C}omputational {L}inguistics}.

\bibitem[{Iannone et~al.(2007)Iannone, Palmisano, and Fanizzi}]{DBLP:journals/apin/IannonePF07}
Luigi Iannone, Ignazio Palmisano, and Nicola Fanizzi. 2007.
\newblock \href {https://doi.org/10.1007/s10489-006-0011-5} {An algorithm based on counterfactuals for concept learning in the semantic web}.
\newblock \emph{Appl. Intell.}, 26(2):139--159.

\bibitem[{Jackermeier et~al.(2023)Jackermeier, Chen, and Horrocks}]{DBLP:journals/corr/abs-2301-11118}
Mathias Jackermeier, Jiaoyan Chen, and Ian Horrocks. 2023.
\newblock \href {https://doi.org/10.48550/arXiv.2301.11118} {Box2el: Concept and role box embeddings for the description logic {EL++}}.
\newblock \emph{CoRR}, abs/2301.11118.

\bibitem[{Jurgens and Pilehvar(2016)}]{jurgens-pilehvar-2016-semeval}
David Jurgens and Mohammad~Taher Pilehvar. 2016.
\newblock \href {https://doi.org/10.18653/v1/S16-1169} {{S}em{E}val-2016 task 14: Semantic taxonomy enrichment}.
\newblock In \emph{Proceedings of the 10th International Workshop on Semantic Evaluation ({S}em{E}val-2016)}, pages 1092--1102, San Diego, California. Association for Computational Linguistics.

\bibitem[{Kipf and Welling(2017)}]{DBLP:conf/iclr/KipfW17}
Thomas~N. Kipf and Max Welling. 2017.
\newblock \href {https://openreview.net/forum?id=SJU4ayYgl} {Semi-supervised classification with graph convolutional networks}.
\newblock In \emph{5th International Conference on Learning Representations, {ICLR} 2017, Toulon, France, April 24-26, 2017, Conference Track Proceedings}. OpenReview.net.

\bibitem[{Kozareva and Hovy(2010)}]{kozareva-hovy-2010-semi}
Zornitsa Kozareva and Eduard Hovy. 2010.
\newblock \href {https://aclanthology.org/D10-1108} {A semi-supervised method to learn and construct taxonomies using the web}.
\newblock In \emph{Proceedings of the 2010 Conference on Empirical Methods in Natural Language Processing}, pages 1110--1118, Cambridge, MA. Association for Computational Linguistics.

\bibitem[{Kulmanov et~al.(2019)Kulmanov, Liu{-}Wei, Yan, and Hoehndorf}]{DBLP:conf/ijcai/KulmanovLYH19}
Maxat Kulmanov, Wang Liu{-}Wei, Yuan Yan, and Robert Hoehndorf. 2019.
\newblock \href {https://doi.org/10.24963/ijcai.2019/845} {{EL} embeddings: Geometric construction of models for the description logic {EL++}}.
\newblock In \emph{Proceedings of the Twenty-Eighth International Joint Conference on Artificial Intelligence, {IJCAI} 2019, Macao, China, August 10-16, 2019}, pages 6103--6109. ijcai.org.

\bibitem[{Le et~al.(2019)Le, Roller, Papaxanthos, Kiela, and Nickel}]{le-etal-2019-inferring}
Matthew Le, Stephen Roller, Laetitia Papaxanthos, Douwe Kiela, and Maximilian Nickel. 2019.
\newblock \href {https://doi.org/10.18653/v1/P19-1313} {Inferring concept hierarchies from text corpora via hyperbolic embeddings}.
\newblock In \emph{Proceedings of the 57th Annual Meeting of the Association for Computational Linguistics}, pages 3231--3241, Florence, Italy. Association for Computational Linguistics.

\bibitem[{Li et~al.(2021)Li, Bouraoui, Camacho{-}Collados, Anke, Gu, and Schockaert}]{DBLP:conf/ijcai/LiBCAGS21}
Na~Li, Zied Bouraoui, Jos{\'{e}} Camacho{-}Collados, Luis~Espinosa Anke, Qing Gu, and Steven Schockaert. 2021.
\newblock \href {https://doi.org/10.24963/ijcai.2021/530} {Modelling general properties of nouns by selectively averaging contextualised embeddings}.
\newblock In \emph{Proceedings of the Thirtieth International Joint Conference on Artificial Intelligence, {IJCAI} 2021, Virtual Event / Montreal, Canada, 19-27 August 2021}, pages 3850--3856. ijcai.org.

\bibitem[{Li et~al.(2019)Li, Bouraoui, and Schockaert}]{DBLP:conf/semweb/LiBS19}
Na~Li, Zied Bouraoui, and Steven Schockaert. 2019.
\newblock \href {https://doi.org/10.1007/978-3-030-30793-6\_25} {Ontology completion using graph convolutional networks}.
\newblock In \emph{The Semantic Web - {ISWC} 2019 - 18th International Semantic Web Conference, Auckland, New Zealand, October 26-30, 2019, Proceedings, Part {I}}, volume 11778 of \emph{Lecture Notes in Computer Science}, pages 435--452. Springer.

\bibitem[{Li et~al.(2023)Li, Kteich, Bouraoui, and Schockaert}]{DBLP:conf/sigir/LiKBS23}
Na~Li, Hanane Kteich, Zied Bouraoui, and Steven Schockaert. 2023.
\newblock \href {https://doi.org/10.1145/3539618.3591667} {Distilling semantic concept embeddings from contrastively fine-tuned language models}.
\newblock In \emph{Proceedings of the 46th International {ACM} {SIGIR} Conference on Research and Development in Information Retrieval, {SIGIR} 2023, Taipei, Taiwan, July 23-27, 2023}, pages 216--226. {ACM}.

\bibitem[{Liu et~al.(2021{\natexlab{a}})Liu, Vuli{\'c}, Korhonen, and Collier}]{liu-etal-2021-fast}
Fangyu Liu, Ivan Vuli{\'c}, Anna Korhonen, and Nigel Collier. 2021{\natexlab{a}}.
\newblock \href {https://doi.org/10.18653/v1/2021.emnlp-main.109} {Fast, effective, and self-supervised: Transforming masked language models into universal lexical and sentence encoders}.
\newblock In \emph{Proceedings of the 2021 Conference on Empirical Methods in Natural Language Processing}, pages 1442--1459, Online and Punta Cana, Dominican Republic. Association for Computational Linguistics.

\bibitem[{Liu et~al.(2021{\natexlab{b}})Liu, Liu, Collier, Korhonen, and Vuli{\'c}}]{liu-etal-2021-mirrorwic}
Qianchu Liu, Fangyu Liu, Nigel Collier, Anna Korhonen, and Ivan Vuli{\'c}. 2021{\natexlab{b}}.
\newblock \href {https://doi.org/10.18653/v1/2021.conll-1.44} {{M}irror{W}i{C}: On eliciting word-in-context representations from pretrained language models}.
\newblock In \emph{Proceedings of the 25th Conference on Computational Natural Language Learning}, pages 562--574, Online. Association for Computational Linguistics.

\bibitem[{Liu et~al.(2019)Liu, Ott, Goyal, Du, Joshi, Chen, Levy, Lewis, Zettlemoyer, and Stoyanov}]{DBLP:journals/corr/abs-1907-11692}
Yinhan Liu, Myle Ott, Naman Goyal, Jingfei Du, Mandar Joshi, Danqi Chen, Omer Levy, Mike Lewis, Luke Zettlemoyer, and Veselin Stoyanov. 2019.
\newblock {RoBERTa}: {A} robustly optimized {BERT} pretraining approach.
\newblock \emph{CoRR}, abs/1907.11692.

\bibitem[{Mikolov et~al.(2013)Mikolov, Yih, and Zweig}]{mikolov-etal-2013-linguistic}
Tomas Mikolov, Wen-tau Yih, and Geoffrey Zweig. 2013.
\newblock \href {https://aclanthology.org/N13-1090} {Linguistic regularities in continuous space word representations}.
\newblock In \emph{Proceedings of the 2013 Conference of the North {A}merican Chapter of the Association for Computational Linguistics: Human Language Technologies}, pages 746--751, Atlanta, Georgia. Association for Computational Linguistics.

\bibitem[{Mondal et~al.(2021)Mondal, Bhatia, and Mutharaju}]{DBLP:conf/aaaiss/MondalBM21}
Sutapa Mondal, Sumit Bhatia, and Raghava Mutharaju. 2021.
\newblock \href {https://ceur-ws.org/Vol-2846/paper19.pdf} {Emel++: Embeddings for {EL++} description logic}.
\newblock In \emph{Proceedings of the {AAAI} 2021 Spring Symposium on Combining Machine Learning and Knowledge Engineering {(AAAI-MAKE} 2021), Stanford University, Palo Alto, California, USA, March 22-24, 2021}, volume 2846 of \emph{{CEUR} Workshop Proceedings}. CEUR-WS.org.

\bibitem[{Nickel and Kiela(2017)}]{DBLP:conf/nips/NickelK17}
Maximilian Nickel and Douwe Kiela. 2017.
\newblock \href {https://proceedings.neurips.cc/paper/2017/hash/59dfa2df42d9e3d41f5b02bfc32229dd-Abstract.html} {Poincar{\'{e}} embeddings for learning hierarchical representations}.
\newblock In \emph{Advances in Neural Information Processing Systems 30: Annual Conference on Neural Information Processing Systems 2017, December 4-9, 2017, Long Beach, CA, {USA}}, pages 6338--6347.

\bibitem[{Ozaki(2020)}]{DBLP:journals/ki/Ozaki20}
Ana Ozaki. 2020.
\newblock \href {https://doi.org/10.1007/s13218-020-00656-9} {Learning description logic ontologies: Five approaches. where do they stand?}
\newblock \emph{K{\"{u}}nstliche Intell.}, 34(3):317--327.

\bibitem[{Peng et~al.(2022)Peng, Tang, Kulmanov, Niu, and Hoehndorf}]{DBLP:journals/corr/abs-2202-14018}
Xi~Peng, Zhenwei Tang, Maxat Kulmanov, Kexin Niu, and Robert Hoehndorf. 2022.
\newblock \href {http://arxiv.org/abs/2202.14018} {Description logic {EL++} embeddings with intersectional closure}.
\newblock \emph{CoRR}, abs/2202.14018.

\bibitem[{Pennington et~al.(2014)Pennington, Socher, and Manning}]{pennington-etal-2014-glove}
Jeffrey Pennington, Richard Socher, and Christopher Manning. 2014.
\newblock \href {https://doi.org/10.3115/v1/D14-1162} {{G}lo{V}e: Global vectors for word representation}.
\newblock In \emph{Proceedings of the 2014 Conference on Empirical Methods in Natural Language Processing ({EMNLP})}, pages 1532--1543, Doha, Qatar. Association for Computational Linguistics.

\bibitem[{Roller et~al.(2014)Roller, Erk, and Boleda}]{roller-etal-2014-inclusive}
Stephen Roller, Katrin Erk, and Gemma Boleda. 2014.
\newblock \href {https://aclanthology.org/C14-1097} {Inclusive yet selective: Supervised distributional hypernymy detection}.
\newblock In \emph{Proceedings of {COLING} 2014, the 25th International Conference on Computational Linguistics: Technical Papers}, pages 1025--1036, Dublin, Ireland. Dublin City University and Association for Computational Linguistics.

\bibitem[{Sarker and Hitzler(2019)}]{DBLP:conf/aaai/SarkerH19}
Md.~Kamruzzaman Sarker and Pascal Hitzler. 2019.
\newblock \href {https://doi.org/10.1609/aaai.v33i01.33013036} {Efficient concept induction for description logics}.
\newblock In \emph{The Thirty-Third {AAAI} Conference on Artificial Intelligence, {AAAI} 2019, The Thirty-First Innovative Applications of Artificial Intelligence Conference, {IAAI} 2019, The Ninth {AAAI} Symposium on Educational Advances in Artificial Intelligence, {EAAI} 2019, Honolulu, Hawaii, USA, January 27 - February 1, 2019}, pages 3036--3043. {AAAI} Press.

\bibitem[{Schlichtkrull et~al.(2018)Schlichtkrull, Kipf, Bloem, van~den Berg, Titov, and Welling}]{DBLP:conf/esws/SchlichtkrullKB18}
Michael~Sejr Schlichtkrull, Thomas~N. Kipf, Peter Bloem, Rianne van~den Berg, Ivan Titov, and Max Welling. 2018.
\newblock \href {https://doi.org/10.1007/978-3-319-93417-4\_38} {Modeling relational data with graph convolutional networks}.
\newblock In \emph{The Semantic Web - 15th International Conference, {ESWC} 2018, Heraklion, Crete, Greece, June 3-7, 2018, Proceedings}, volume 10843 of \emph{Lecture Notes in Computer Science}, pages 593--607. Springer.

\bibitem[{Shang et~al.(2020)Shang, Dash, Chowdhury, Mihindukulasooriya, and Gliozzo}]{shang-etal-2020-taxonomy}
Chao Shang, Sarthak Dash, Md. Faisal~Mahbub Chowdhury, Nandana Mihindukulasooriya, and Alfio Gliozzo. 2020.
\newblock \href {https://doi.org/10.18653/v1/2020.acl-main.199} {Taxonomy construction of unseen domains via graph-based cross-domain knowledge transfer}.
\newblock In \emph{Proceedings of the 58th Annual Meeting of the Association for Computational Linguistics}, pages 2198--2208, Online. Association for Computational Linguistics.

\bibitem[{Shen et~al.(2020)Shen, Shen, Xiong, Wang, Wang, and Han}]{DBLP:conf/www/ShenSX0W020}
Jiaming Shen, Zhihong Shen, Chenyan Xiong, Chi Wang, Kuansan Wang, and Jiawei Han. 2020.
\newblock \href {https://doi.org/10.1145/3366423.3380132} {{TaxoExpan}: Self-supervised taxonomy expansion with position-enhanced graph neural network}.
\newblock In \emph{{WWW} '20: The Web Conference 2020, Taipei, Taiwan, April 20-24, 2020}, pages 486--497. {ACM} / {IW3C2}.

\bibitem[{Smaili et~al.(2018)Smaili, Gao, and Hoehndorf}]{DBLP:journals/bioinformatics/SmailiGH18}
Fatima~Zohra Smaili, Xin Gao, and Robert Hoehndorf. 2018.
\newblock \href {https://doi.org/10.1093/bioinformatics/bty259} {Onto2vec: joint vector-based representation of biological entities and their ontology-based annotations}.
\newblock \emph{Bioinform.}, 34(13):i52--i60.

\bibitem[{Smaili et~al.(2019)Smaili, Gao, and Hoehndorf}]{DBLP:journals/bioinformatics/SmailiGH19}
Fatima~Zohra Smaili, Xin Gao, and Robert Hoehndorf. 2019.
\newblock \href {https://doi.org/10.1093/bioinformatics/bty933} {Opa2vec: combining formal and informal content of biomedical ontologies to improve similarity-based prediction}.
\newblock \emph{Bioinform.}, 35(12):2133--2140.

\bibitem[{Speer et~al.(2017)Speer, Chin, and Havasi}]{DBLP:conf/aaai/SpeerCH17}
Robyn Speer, Joshua Chin, and Catherine Havasi. 2017.
\newblock \href {https://doi.org/10.1609/AAAI.V31I1.11164} {Conceptnet 5.5: An open multilingual graph of general knowledge}.
\newblock In \emph{Proceedings of the Thirty-First {AAAI} Conference on Artificial Intelligence, February 4-9, 2017, San Francisco, California, {USA}}, pages 4444--4451. {AAAI} Press.

\bibitem[{Takeoka et~al.(2021)Takeoka, Akimoto, and Oyamada}]{takeoka-etal-2021-low}
Kunihiro Takeoka, Kosuke Akimoto, and Masafumi Oyamada. 2021.
\newblock \href {https://doi.org/10.18653/v1/2021.emnlp-main.217} {Low-resource taxonomy enrichment with pretrained language models}.
\newblock In \emph{Proceedings of the 2021 Conference on Empirical Methods in Natural Language Processing}, pages 2747--2758, Online and Punta Cana, Dominican Republic. Association for Computational Linguistics.

\bibitem[{Velickovic et~al.(2018)Velickovic, Cucurull, Casanova, Romero, Li{\`{o}}, and Bengio}]{DBLP:conf/iclr/VelickovicCCRLB18}
Petar Velickovic, Guillem Cucurull, Arantxa Casanova, Adriana Romero, Pietro Li{\`{o}}, and Yoshua Bengio. 2018.
\newblock \href {https://openreview.net/forum?id=rJXMpikCZ} {Graph attention networks}.
\newblock In \emph{6th International Conference on Learning Representations, {ICLR} 2018, Vancouver, BC, Canada, April 30 - May 3, 2018, Conference Track Proceedings}. OpenReview.net.

\bibitem[{Vilnis et~al.(2018)Vilnis, Li, Murty, and McCallum}]{vilnis-etal-2018-probabilistic}
Luke Vilnis, Xiang Li, Shikhar Murty, and Andrew McCallum. 2018.
\newblock \href {https://doi.org/10.18653/v1/P18-1025} {Probabilistic embedding of knowledge graphs with box lattice measures}.
\newblock In \emph{Proceedings of the 56th Annual Meeting of the Association for Computational Linguistics (Volume 1: Long Papers)}, pages 263--272, Melbourne, Australia. Association for Computational Linguistics.

\bibitem[{Xiong et~al.(2022)Xiong, Potyka, Tran, Nayyeri, and Staab}]{DBLP:conf/semweb/XiongPTNS22}
Bo~Xiong, Nico Potyka, Trung{-}Kien Tran, Mojtaba Nayyeri, and Steffen Staab. 2022.
\newblock \href {https://doi.org/10.1007/978-3-031-19433-7\_2} {Faithful embeddings for {EL++} knowledge bases}.
\newblock In \emph{The Semantic Web - {ISWC} 2022 - 21st International Semantic Web Conference, Virtual Event, October 23-27, 2022, Proceedings}, volume 13489 of \emph{Lecture Notes in Computer Science}, pages 22--38. Springer.

\bibitem[{Yang et~al.(2015)Yang, Yih, He, Gao, and Deng}]{DBLP:journals/corr/YangYHGD14a}
Bishan Yang, Wen{-}tau Yih, Xiaodong He, Jianfeng Gao, and Li~Deng. 2015.
\newblock \href {http://arxiv.org/abs/1412.6575} {Embedding entities and relations for learning and inference in knowledge bases}.
\newblock In \emph{3rd International Conference on Learning Representations, {ICLR} 2015, San Diego, CA, USA, May 7-9, 2015, Conference Track Proceedings}.

\end{thebibliography}

\appendix
\begin{table*}
\footnotesize
\centering
\begin{tabular}{ll}
\toprule
\textbf{Model Name} & \textbf{URL} \\
\midrule
Llama2-7B  & 
\url{https://huggingface.co/meta-llama/Llama-2-7b-hf} \\
Llama2-7B-Chat & 
\url{https://huggingface.co/meta-llama/Llama-2-7b-chat-hf} \\
Llama2-13B  & 
\url{https://huggingface.co/meta-llama/Llama-2-13b-hf} \\
Llama2-13B-Chat & 
\url{https://huggingface.co/meta-llama/Llama-2-13b-chat-hf} \\
Llama2-7B-32K-Instruct & 
\url{https://huggingface.co/togethercomputer/Llama-2-7B-32K-Instruct} \\
Mistral-7B  & 
\url{https://huggingface.co/mistralai/Mistral-7B-v0.1} \\
Mistral-7B-Instruct  & 
\url{https://huggingface.co/mistralai/Mistral-7B-Instruct-v0.2} \\
Vicuna-13B & 
\url{https://huggingface.co/lmsys/vicuna-13b-v1.5 } \\
Vicuna-13B-16K & 
\url{https://huggingface.co/lmsys/vicuna-13b-v1.5-16k}\\
\bottomrule
\end{tabular}
\caption{Specification of the LLMs that were used in our experiments.\label{tabDetailsLLMs}}
\end{table*}

\section{Experimental Details}

\paragraph{Considered Models}
Table \ref{tabDetailsLLMs} gives an overview of the language models that were used in our experiments, together with information about where they can be obtained.

\paragraph{Training Details}
We use the rule-based verbalizer provided by the DeepOnto library to convert the rules into textual inputs. For instance, the simple concept $\mathsf{RedWine}$ is converted to the  term ``red wine'', while the concept $\exists \mathsf{hasColor}.\mathsf{Red}$ is converted to the phrase ``something that has color red''. 
For training with DeepOnto, we set the learning rate to 1e-5, weight decay to 1e-2, the number of epochs to 3, the batch size of the training and development sets to 8, and the batch size of test sets to 16.  
For tuning the GNN models, we select the number of layers from \{2, 3, 4, 5\}. For GAT and GATv2, we select the number of attention heads from \{4, 8, 16\} and fix the negative slope of the LeakyReLU activations as 0.2. In all GNN models, we use dropout to avoid over-fitting. For GAT and GATv2, the dropout rate of attention layers is set to 0.2. For all GNN models, the dropout rate of non-attention layers is set to 0.5. We select the dimension of the hidden layers from \{8, 16, 32,64\}.
All GNN models are optimised using AdamW, with a learning rate of 1e-2 and weight decay of 5e-2. We train the models for 200 epochs and select the best checkpoint based on the validation split.
To fine-tune the LLMs, we rely on QLoRA, which combines 4-bit quantization via BitsAndBytes with Low-Rank Adaptation (LoRA) to enable efficient model optimization. 

\section{Additional Analysis}

\begin{table*}[t]
\centering
\footnotesize
\begin{tabular}{lcccccccc}
\toprule
& \textbf{Wine} & \textbf{Economy} & \textbf{Olympics} & \textbf{Transport} & \textbf{SUMO} & \textbf{FoodOn} & \textbf{GO} & \textbf{Average} \\
\midrule
Llama2-7B  & 45.0 & 63.3 & 55.1 & 60.0 & 64.0 & 75.1 & 72.7 & 62.2\\
Llama2-7B-Chat & 50.8 & 56.0 & 50.2 & 53.5 & 60.6 & 74.5 & 69.0 & 59.2\\
Llama2-13B  & 52.2 & 68.2 & 55.6 & 62.0 & 69.5 & 78.2 & 77.2 & 66.1\\
Llama2-13B-Chat & 54.4 & 66.0 & 53.6 & 55.0 & 70.1 & 76.8 & 75.6 & 64.5\\
Llama2-7B-32K-Instruct & 45.8 & 66.2 & 64.5 & 60.4 & 69.0 & 75.7 & 70.1 & 64.5\\
Mistral-7B  & 62.3 & 71.9 & 66.5 & 70.3 & 72.5 & 77.5 & 75.1 & 70.9\\
Mistral-7B-Instruct  & 62.3 & 70.7 & 64.5 & 70.0 & 71.7 & 77.0 & 76.8 & 70.4\\
Vicuna-13B & 54.1 & 78.4 & 73.0 & 69.4 & 72.5 & 77.3 & 76.8 & 71.6\\
Vicuna-13B-16K & 57.6 & 75.7 & 74.6 & 69.4 & 72.1 & 76.9 & 74.5 & 71.5\\
\bottomrule
\end{tabular}
\caption{Comparison of different variants of the considered LLMs. \label{tabMainResultsVariantAnalysis}}
\end{table*}

\begin{table*}[t]
\centering
\footnotesize
\begin{tabular}{lcccccccc}
\toprule
& \textbf{Wine} & \textbf{Economy} & \textbf{Olympics} & \textbf{Transport} & \textbf{SUMO} & \textbf{FoodOn} & \textbf{GO} & \textbf{Average}\\
\midrule
Llama2-7B  & 53.8 & 71.4 & 71.45 & 60.2 & 69.9 & 75.3 & 73.4 & 67.9\\
Llama2-7B-Chat & 37.1 & 68.2 & 66.7 & 59.7 & 68.0 & 73.5 & 72.3 & 63.7\\
Llama2-13B  & 45.6 & 69.8 & 68.1 & 58.5 & 68.2 & 75.3 & 75.5 & 65.9\\
Llama2-13B-Chat & 48.3 & 71.9 & 63.3 & 60.8 & 67.7 & 74.3 & 74.4 & 65.8\\
Llama2-7B-32K-Instruct & 51.3 & 65.0 & 74.6 & 58.2 & 65.8 & 72.2 & 69.3 & 65.2\\
Mistral-7B  & 58.6 & 78.9 & 66.7 & 60.0 & 71.3 & 79.1 & 77.1 & 70.2\\
Mistral-7B-Instruct  & 57.7 & 70.2 & 71.4 & 59.5 & 69.8 & 76.2 & 75.4 & 68.6\\
Vicuna-13B & 48.6 & 72.3 & 63.5 & 59.7 & 71.8 & 77.1 & 76.3 & 67.0\\
Vicuna-13B-16K & 54.6 & 71.8 & 71.4 & 58.8 & 70.5 & 75.5 & 76.0 & 68.4\\
\bottomrule
\end{tabular}
\caption{Results for the NLI models when jointly trained on all ontologies. \label{tabExperimentJointlyTrained}}
\end{table*}

\begin{table*}[t]
\centering
\footnotesize
\begin{tabular}{lccccc}
\toprule
& \textbf{Prompt 1} & \textbf{Prompt 2} & \textbf{Prompt 3} & \textbf{Prompt 4} & \textbf{Prompt 5} \\
\midrule
Llama2-7B  & 66.4 & 61.9 & 62.9 & 67.2 & 67.1\\
Llama2-7B-Chat & 61.3 & 63.0 & 63.7 & 65.0 & 64.8\\
Llama2-13B  & 69.5 & 69.9 & 71.2 & 70.4 & 70.9\\
Llama2-13B-Chat & 65.4 & 65.4 & 63.3 & 65.2 & 65.8\\
Llama2-7B-32K-Instruct & 67.0 & 67.4 & 67.8 & 63.2 & 65.8\\
Mistral-7B  & 72.1 & 72.3 & 71.7 & 71.9 & 72.3\\
Mistral-7B-Instruct  & 71.9 & 71.3 & 70.9 & 70.6 & 69.7\\
Vicuna-13B & 69.7 & 69.0 & 68.4 & 69.8 & 68.6 \\
Vicuna-13B-16K & 69.8 & 69.0 & 69.8 & 68.2 & 70.0\\
\bottomrule
\end{tabular}
\caption{Analysis of the performance of different prompts in terms of F1 ($\%$) on a combined test set containing 100 examples from each of the ontologies. \label{tabPromptAnalysis}}
\end{table*}

\begin{table}[t]
\centering
\footnotesize
\begin{tabular}{lcc}
\toprule
          & \textbf{DistMult} & \textbf{TransE} \\ 
          \midrule
Wine      & 72.2     & 70.8   \\ 
Economy   & 14.6     & 15.7   \\ 
Olympics  & 27.9     & 27.3   \\ 
Transport & 47.0     & 45.9  \\
\bottomrule
\end{tabular}
\caption{Comparison between DistMult and TransE as scoring function for the Binary Template model (F1\%).}
\label{transe_distmult}
\end{table}

\paragraph{Variants of LLMs}
In the main experiments, we used the base versions of Llama 2, Mistral and Vicuna. 
Table \ref{tabMainResultsVariantAnalysis} compares these models with a number of variants:  chat and instruction fine-tuned versions of Llama 2, the instruction fine-tuned version of Mistral, and a variant of Vicuna with a larger context window. As we can see, these variants generally perform slightly worse than the base models (with the exception that the instruction fine-tuned Llama2-7B model outperforms the base variant).

\paragraph{Joint Training}
In the main experiments (Table \ref{tabMainResults}), the NLI based models were trained on each ontology separately. This has the advantage that the resulting models are specialised towards the given ontology, which can be important if ontologies use concepts in idiosyncratic ways, among others. However, jointly training these models on all ontologies together also has some possible advantages. For instance, some of the smaller ontologies may not have enough examples to enable successful fine-tuning of LLMs. Moreover, the models might generalise better by being exposed to a more diverse set of training examples. Table \ref{tabExperimentJointlyTrained} shows the results we obtained with this joint training strategy. We can see that this leads to worse results for the best-performing models. However, some configurations, such as the 7B parameter Llama 2 models, benefit from this joint training strategy.

\paragraph{Scoring Function}
For the binary template model, in our main experiments we have relied on a bilinear scoring function. Another possibility, inspired by TransE, is to use the following:
$$
\textit{conf}(\rho,X,Y) = \sigma(\|(\mathbf{y} - \mathbf{x} -\mathbf{a}_{\rho}\|_2 - b_{\rho})
$$
with $\mathbf{a}_{\rho}\in\mathbb{R}^n$ and $b\in\mathbb{R}$. A comparison between both alternatives is shown in Table \ref{transe_distmult}. As can be seen, the results are broadly similar.

\paragraph{Prompt Analysis}
In Table \ref{tabPromptAnalysis}, we compare the performance of different prompts. For this analysis, we use a test set that consists of 100 examples from each of the seven ontologies (50 positive and 50 negative examples) 
The prompts that we considered are as follows:
\begin{itemize}
\item Prompt 1: Classify the text into True or False. Reply with only one word: True or False. Determine if the following statement is valid:
\item Prompt 2: Assess the validity of the following statement. Reply with only one word: True or False. Determine if the following statement is valid:
\item Prompt 3: Assess the validity of the following rule. Reply with only one word: True or False. Determine if the following rule is valid:
\item Prompt 4: Classify the text into True or False. Reply with only one word: True or False. Determine if the following is a valid rule:
\item Prompt 5: Classify the text into True or False. Reply with only one word: True or False. Determine if the following is valid statement:
\end{itemize}
In all cases, the prompt is followed by a statement of the form \textit{\textsc{Rule Body} implies \textsc{Rule Head}}. As we can see in Table Table \ref{tabPromptAnalysis}, the performance of these prompts is comparable.

\paragraph{Qualitative Analysis}
Below are examples of rules that were identified by the GCN with unary templates but not by the Llama2-13B NLI model:
\begin{itemize}
\item Beaujolais implies something that has sugar Dry 
\item Chenin Blanc implies something that has flavor Moderate
\item Avocado implies Grocery Produce 
\item Ready-To-Eat Bakery Product implies Bakery Food Product 
\item Smoked and Frozen Cod Fillet implies Cod Fillet 
\item Rings implies Artistic Gymnastics 
\item Platform implies Diving 
\item LCAC implies Military Vehicle 
\item Abort implies Computer Process 
\item Food Distribution Operation implies Military Operation 
\item Head End Car implies Railcar 
\item Petite Syrah implies something that has sugar Dry 
\item Pauillac implies something that has body Full 
\item Team Event and Individual Event implies Contradiction 
\item Railroad Track and Bulkhead implies Contradiction 
\item Human Habitation Artifact and Ship Deck implies Contradiction
\end{itemize}
Below are examples of rules that were identified by the Llama2-13B NLI model but not by the GCN with unary templates:
\begin{itemize}
\item Sauternes implies something located in Sauterne Region
\item Muscadet implies something made from grape Pinot Blanc Grape
\item Chianti implies something located in Chianti Region
\item Fire Boat implies Emergency Vehicle
\item Canal System implies Water Transportation System
\item Radio Navigation Beacon implies Aid To Navigation
\item Machine implies Machinery
\item War implies Violent Contest
\item Telegraph implies Electric Device
\item Womens Team implies something that has member Woman
\item Artistic Gymnastics implies Gymnastics
\item Summer Games implies Olympic Games
\item Cocaine implies Narcotic
\item Plastic implies Manufactured Product
\item Coffee Bean implies Plant Agricultural Product
\end{itemize}

\end{document}